\documentclass[10pt,journal,compsoc]{IEEEtran}
\usepackage[utf8]{inputenc}
\usepackage[T1]{fontenc}  
\usepackage[greek,english]{babel}
\usepackage{alphabeta}
\usepackage{amsmath}
\usepackage{amssymb}
\usepackage{graphicx}
\usepackage{array}
\usepackage{verbatim}
\usepackage{tabularx}
\usepackage{hyperref}
\usepackage{mathtools}

\usepackage{footnote}
\usepackage{makecell}
\makesavenoteenv{tabular}
\usepackage{algorithm}
\usepackage{algpseudocode}
\usepackage{caption}
\usepackage{subcaption}
\usepackage{orcidlink}

\DeclareMathOperator*{\argmax}{arg\,max}
\DeclareMathOperator*{\argmin}{arg\,min}

\hypersetup{
	colorlinks=true,
	linkcolor=blue,
	filecolor=magenta,
	urlcolor=blue,
}

\begin{document}

\title{Dendritic Self-Organizing Maps for Continual Learning}

\author{Kosmas Pinitas\orcidlink{0000-0003-0938-682X},
        Spyridon Chavlis\orcidlink{0000-0002-1046-1201},
        and Panayiota Poirazi\orcidlink{0000-0001-6152-595X}\IEEEauthorrefmark{2}
\IEEEcompsocitemizethanks{\IEEEcompsocthanksitem K. Pinitas was with the  School of Electrical \& Computer Engineering, Technical University of Crete, Greece.\protect\\
E-mail: \href{mailto:kpinitas@isc.tuc.gr}{kpinitas@isc.tuc.gr}
\IEEEcompsocthanksitem S. Chavlis and P. Poirazi are with the Institute of Molecular Biology and Biotechnology, Foundation for Research and Technology-Hellas, Greece.\protect\\
E-mail: \href{mailto:schavlis@imbb.forth.gr}{schavlis@imbb.forth.gr}, \href{mailto:poirazi@imbb.forth.gr}{poirazi@imbb.forth.gr}
\IEEEcompsocthanksitem \IEEEauthorrefmark{2}: Corresponding author}
\thanks{Manuscript received October 2020}}

\IEEEtitleabstractindextext{%
\begin{abstract}
Current deep learning architectures show remarkable performance when trained in large-scale, controlled datasets. However, the predictive ability of these architectures significantly decreases when learning new classes incrementally. This is due to their inclination to forget the knowledge acquired from previously seen data, a phenomenon termed catastrophic-forgetting. On the other hand, Self-Organizing Maps (SOMs) can model the input space utilizing constrained k-means and thus maintain past knowledge. Here, we propose a novel algorithm inspired by biological neurons, termed Dendritic-Self-Organizing Map (DendSOM). DendSOM consists of a single layer of SOMs, which extract patterns from specific regions of the input space accompanied by a set of hit matrices, one per SOM, which estimate the association between units and labels. The best-matching unit of an input pattern is selected using the maximum cosine similarity rule, while the point-wise mutual information is employed for class inference. DendSOM performs unsupervised feature extraction as it does not use labels for targeted updating of the weights. It outperforms classical SOMs and several state-of-the-art continual learning algorithms on benchmark datasets, such as the Split-MNIST and Split-CIFAR-10. We propose that the incorporation of neuronal properties in SOMs may help remedy catastrophic forgetting.
\end{abstract}

}

\maketitle

\section{Introduction}
\IEEEPARstart{D}{eep} Learning (DL) architectures have achieved state-of-the-art performance in several fields such as speech recognition \cite{kiela2020hateful}, computer vision \cite{borji2014human}, and disease diagnosis \cite{shen2019artificial}. However, they are not as successful for learning multiple tasks sequentially \cite{awasthi2019continual}. Most deep neural networks (DNNs) can only be trained when the entire labeled dataset is at one's disposal. Re-training is often required when new data becomes available, due to catastrophic-forgetting. On the other hand, by leveraging their past experiences, humans can adapt to environments where new tasks emerge continuously and achieve impressive performance on real-world problems. Consequently, humans can continuously acquire new knowledge without forgetting previously learned information, which is not the case for DNNs.

While extensively applied to numerous fields, DNNs remain vaguely inspired by biological neural networks since the computational unit (node) of a DNN is a simplistic model of a biological neuron. However, unlike typical nodes used in DNNs, biological neurons have dendrites: thin processes that are specialized for receiving and locally transforming information from other neurons \cite{stuart2016dendrites}. Incoming signals arrive in the form of synaptic activity, the effect of which is determined by the weight of the activated inputs and their ability to open voltage-gated ionic mechanisms that lie across the neuronal membrane. The opening of such mechanisms results in the nonlinear transformation of the original signal. As a consequence, input at a given synapse can affect its neighboring synapses in nonlinear ways. Moreover, unlike typical DNNs, biological networks are not fully connected. Each neuron receives synaptic input that projects on its dendrites solely and not from the dendrites of other neurons, thus creating a sparsely connected, tree-like network structure. In such biological networks, each neuron samples the environment in a restricted manner - through its dendrites.

Notably, biological neural networks can change throughout life, a process known as plasticity. This plasticity is not limited to weight updates, as in most DNNs. It can also include changes in the spatial organization of synapses, the coupling strength between dendrites and their somata, as well as changes in the nonlinear mechanisms within dendrites \cite{branco2010single,losonczy2008compartmentalized,london2005dendritic,spruston2008pyramidal}. Because of these properties, dendrites are considered the basic computational unit for integrating synaptic input, undergoing synaptic plasticity, and storing multiple complex features of the synaptic input. Thus, the incorporation of dendrites is likely to assist current DL models in obtaining greater information processing power \cite{chavlis2021drawing}, and recent findings (see Section \ref{Related Work}) support this notion.

Drawing inspiration from the structure and properties of dendrites, we introduce a SOM-based architecture that models the dendritic nonlinearities and permits local, parallel computations. The DendSOM map model consists of a single layer of SOMs. Each SOM is treated as a set of dendrites, and its units are connected to a single soma. In particular, SOMs receive and model different subregions of the input space, and the somata estimate the association between the best-matching units and class labels. We test our model on classification and continual learning (CL) scenarios and show that DendSOM can alleviate the effects of catastrophic forgetting. The rest of this paper is organized as follows. In the next section, we provide the essential literature review, and in section 3, we revisit the SOM algorithm. Sections 4 and 5 describe the proposed architecture in detail and present the experimental setup and results.

\section{Related Work} \label{Related Work}
In CL, the term "catastrophic forgetting" refers to the propensity of a model to forget the previously learned tasks when re-trained on a new, unseen task \cite{nguyen2019toward}. Over the last years, several models have managed to deal with this problem to some extent. Context-dependent Gating (XdG) \cite{masse2018alleviating} does so by deactivating a random subset of units on each layer for each task. Therefore, XdG forces other network parts to learn the subsequent tasks, partially avoiding overwriting. However, this method can only be utilized for learning problems in which the task is known since it requires a unique signal for each given task.

The Learning without Forgetting (LwF) algorithm \cite{li2017learning} tries to address catastrophic forgetting by forcing predictive stability. It contains shared and task-specific parameters. The goal is to add task-specific parameters for any new task while learning the shared parameters that work well on all tasks, using only data and labels from the new task. The Deep Generative Replay (DGR) model \cite{shin2017continual}, on the other hand, consists of a classifier and a generator network. The goal of the classifier is to learn a mapping from the input to the output space, while the generator aims to learn the distribution of the data in order to sample data based on previous experience. Both algorithms are trained on the first batch of input data, but when the next batch arrives, the generator produces the replay data mixed with the training data of the current batch, and both models are trained in the new, up-sampled dataset.

Elastic Weight Consolidation (EWC) \cite{kirkpatrick2017overcoming} is a prior-focused regularization-based algorithm which penalizes updates on important weights by adding a regularization term to the loss function. In this method, the critical parameters must remain stable since a slight change can cause a dramatic increase in the loss, due to the high curvature of the loss surface towards their direction. A neuro-inspired algorithm called Synaptic Intelligence (SI) \cite{zenke2017continual}, allows weights (synapses) to estimate their importance on previous tasks and penalizes changes to the most vital weights. SI differs from EWC because it computes the weight importance during stochastic gradient descent (SGD), unlike EWC, which computes the Fisher Information Matrix at the end of the training on the batch.

Copy Weight with Reinit (CWR) \cite{maltoni2019continuous} is considered a baseline technique for learning from sequential batches in each CL scenario. When the task label is given, the algorithm freezes the shared weights $\bar{\Theta}$ after the completion of a task and extends the output layer with new randomly initialized neurons for the new task. However, when the environment can not provide information about the task boundaries due to the unavailability of the task label, the algorithm still follows the same procedure, though, in this case, the architecture is extended for each training batch. This model has two sets of weights for the output layer, the consolidated $cw$ weights that are initialized to zero and used for inference, and the temporary weights $tw$, which are randomly re-initialized before each batch in order to be used for batch training. At the end of the batch training, $tw$ is scaled and copied to $cw$. 

The Copy Weight with Reinit+ (CWR+) \cite{maltoni2019continuous} is a modified version of CWR. The modifications include the global mean-shift of $tw$ and the zero-initialization of the output layer. Additionally, the Architectural and Regularization approach (AR1) \cite{maltoni2019continuous} is a hybrid algorithm that utilizes both parameter isolation and regularization. In particular, AR1 is similar to the CWR+ algorithm since it follows an almost identical training process. However, this algorithm employs SI to introduce the concept of parameter importance and thus tune the shared weights $\bar{\Theta}$ across the entire dataset. Finally, in generative classification \cite{van2021class}, a separate variational autoencoder (VAE) is trained in order to learn the data distribution for each class. Next, importance sampling is employed to estimate the likelihood of a test sample $x$ under the VAE model of class $y$ denoted as $p(x|y)$, while $p(y)$ can either be assumed to follow a normal distribution or is calculated by counting sample observations, where the Bayesian rule is used for inference. Last but not least, a method closer to ours, the Self-Organized Multi-Layer Perceptron (SOMLP) \cite{bashivan2019continual} utilizes a SOM to control the activation of the units on the hidden layer of a Multilayer Perceptron to avoid overwriting of the previously learned information, which leads to catastrophic forgetting.

The main challenge of the algorithms that have alleviated the effects of catastrophic forgetting is that they can only be trained when all classes are known before training. This prerequisite is not realistic in real-world scenarios, where new classes emerge continuously. We propose a novel, semi-unsupervised learning algorithm based on the SOM architecture, which incorporates features discovered in biological dendrites to address this challenge. Specifically, the DendSOM is a shallow network that performs classification and incorporates the following biological properties: i) sparse/restricted sampling of the input space via the use of receptive fields, ii) restricted connectivity through the implementation of soma-specific SOMs (akin to dendrites), iii) local, nonlinear computations mimicking the signal integration properties of biological dendrites and iv) the use of spatially organized memory images (prototypes) to learn information about the input space. We evaluate this architecture on three incremental learning scenarios \cite{van2019three} using the split protocol on the MNIST and CIFAR-10 benchmark datasets. We show that our model performs favorably compared to other state-of-the-art architectures, suggesting that incorporating properties found in real neurons can be beneficial for complex machine learning problems.

\section{Self-Organizing Map}
The SOM is an unsupervised learning algorithm that non-linearly projects high-dimensional input data to a space of lower dimensions (typically two-dimensional). The SOM applies competitive learning to construct the map's topology such that the more similar the input data, the shorter their distance on the map \cite{kohonen2013essentials}.
\subsection{Learning Algorithm} \label{Dend SOM}
The SOM consists of a grid of units. Each unit $i$ has a predetermined position $\mathbf{p_i}$ on the grid, where $\mathbf{p_i} \in \mathbb{N}^{+^m}$ and $m$ is the grid's dimensions. Furthermore, the $i$-th unit maintains a weight vector $\mathbf{w_i} \in \mathbb{R}^k$ where $k$ is the number of dimensions of the input vector $\mathbf{x_n}$ with $n=1,2,\dots, N$, where $N$ denotes the number of samples. The weight vectors are initialized to small random values.

The SOM is trained iteratively, and at each iteration $t$, the learning process performs the following steps:
\begin{enumerate}
	\item Select a sample vector $\mathbf{x}$ randomly from the input dataset $X$
	\item Identify the Best-Matching Unit (BMU) that satisfies:
	\begin{equation}
		BMU = \argmin_{i} ||\mathbf{x}-\mathbf{w}_i||^2_{2} \label{find bmu}
	\end{equation}
	where $||\cdot||_2$ denotes the 2-norm (so-called Euclidean norm).
	\item Update each weight vector $i$ using the following equation:
	\begin{equation}
	\mathbf{w}_i(t+1) = \mathbf{w}_i(t) + a(t)h^{BMU}_{i}(t)\bigg(\mathbf{x}-\mathbf{w_i(t)}\bigg) \label{w_upt}
	\end{equation}

	\noindent
	The learning rate $a(t)$ at training step $t$ is calculated by:
	\begin{equation}
	a(t) = a_0\exp\bigg(-\frac{t}{\lambda}\bigg) \label{lr}
	\end{equation}
	\noindent
	Moreover, the neighborhood function $h^{BMU}_{i}(t)$ at iteration $t$ is calculated as follows:
	\begin{equation}
	h^{BMU}_{i}(t) = \exp\bigg(-\frac{||\mathbf{p}_i-\mathbf{p}_{BMU}||^2_2}{2\sigma(t)}\bigg) \label{h}
	\end{equation}
	\noindent
	where $\mathbf{p}_{BMU}$ denote the position of the BMU on the grid, and $\sigma(t)$ corresponds to the neighborhood radius at step $t$ and it can be calculated by: 
	\begin{equation}
	\sigma(t) = \sigma_0\exp\bigg(-\frac{t}{\lambda}\bigg) \label{sigma}
	\end{equation}
\end{enumerate}

\begin{figure}[ht]
	\centering
	\includegraphics[keepaspectratio,width=0.45\textwidth]{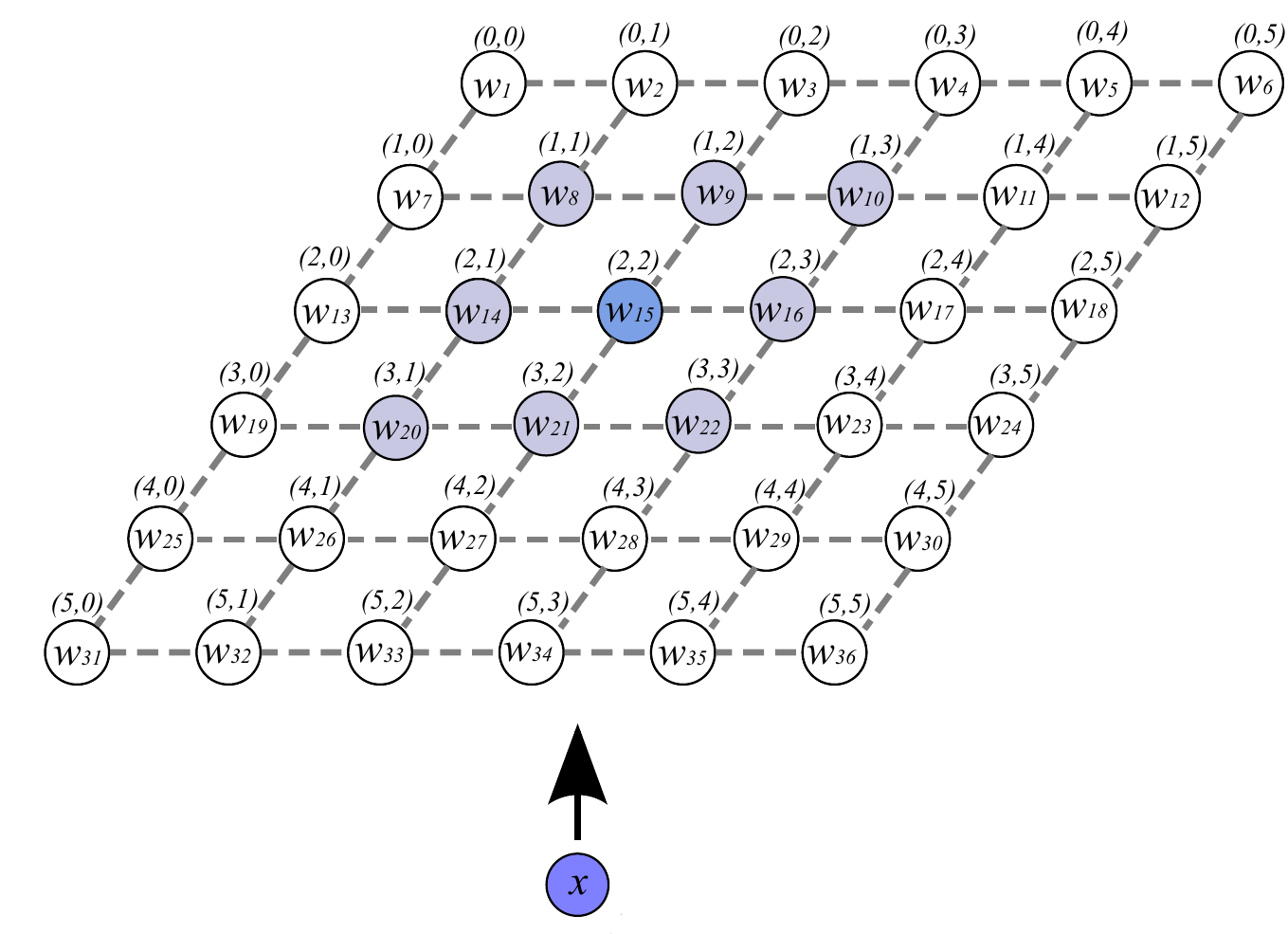}
	\caption{Training process of SOM. In this example, the unit in position $(2,2)$ is selected as the BMU for the input vector $x$.}
	\label{fig:train_som}
\end{figure}

\noindent
During the update step, the BMU and its topological neighbors move closer to the input vector while $a(t)$, $\sigma(t)$ and $h^{BMU}_{i}(t)$ decrease exponentially over time (a schematic representation is depicted in Figure \ref{fig:train_som}. Moreover, $a_0$ and $\sigma_0$ are the initial values of the learning rate and neighborhood radius, respectively, while $\lambda$ is a hyperparameter that controls the decrease rate for both $\sigma(t)$ and $a(t)$.

\section{Dendritic Self-Organizing Map}
Unlike other SOM-based deep learning architectures such as Deep SOM, Deep Convolutional SOM, and Layered-SOM \cite{liu2015deep,aly2020deep,nakagawa2017classification}, the DendSOM proposed here is a shallow neural network that can be trained using unsupervised competitive learning algorithms. Hence, it can be applied to classification and CL tasks. DendSOM employs multiple SOMs to model different subregions of the input space (receptive fields). Like dendrites in biological neurons, each SOM receives spatially-restricted input information and estimates the positional vector of the BMU, which it subsequently propagates to its respective soma, where the association between labels and BMUs is calculated. The incorporation of receptive fields allows for more accurate modeling of the input space since it considers the temporal evolution of the images. However, introducing the notion of receptive fields without restricted connectivity (i.e., computing the BMU solely for the receptive field of each SOM) can significantly increase the training time due to the lack of computational parallelism.
\subsection{Architecture}
As mentioned above, the DendSOM uses multiple SOMs to model the input pattern from different subregions of the input space. Consequently, for an input $\mathbf{x}$ of size $N_1\times N_2$ and SOMs that extract patterns of size $P_1\times P_2$ we need $S_1 \times S_2$ SOMs in order to properly model the input space. Assuming the stride $s_1$, $s_2$ respectively, then $S_1$ and $S_2$ can be calculated by:  
\begin{equation}
    \begin{split}
    S_i = \bigg\lfloor\frac{N_i-P_i}{s_i}\bigg\rfloor+1, \:\:\: i \in \{1,2\}
    \end{split}
\end{equation}

\noindent
where $\lfloor \cdot \rfloor$ denotes the floor operation.
\newline

In our implementation $s_1=s_2$, $N_1 = N_2$, $P_1=P_2$ and thus $S_1=S_2$. 

\subsection{Best Matching Unit}
\begin{figure}[ht]
    \centering
    \includegraphics[keepaspectratio,width=0.45\textwidth]{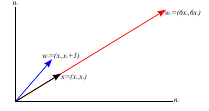}
    \caption{This diagram shows the 2D vectors $x$ (input vector) $w_1$ and $w_2$ (weight vectors). According to the minimum euclidean distance rule, the unit that contains $w_1$ is the best matching unit since $euc(x,w_1)<euc(x,w_2)$. However, the angle between $x$ and $w_2$ is smaller than the the angle between $x$ and $w_1$ and consequently $\cos(x,w_2)>\cos(x,w_1)$. Hence the maximum cosine rule indicates that the best matching unit is the unit with weight vector $w_2$. The cosine similarity rule is a scale invariant rule since it depends only from the inter-pixel relations and not from the exact pixel values.}
    \label{fig:cosine_sim}
\end{figure}

In the original SOM algorithm, the BMU is defined as the unit whose weight vector has the minimum Euclidean distance from the input vector, as shown in equation (\ref{find bmu}). The Euclidean distance corresponds to the L2 norm of a difference between vectors. Thus, the Euclidean distance estimates the similarity between patterns by measuring how close the images are with respect to their pixel intensity. The cosine similarity is a different metric, which is proportional to the dot product of two vectors and inversely proportional to the product of their magnitudes. Patterns with high cosine similarity are located in the same general direction from the origin (Figure \ref{fig:cosine_sim}). In this work, we used the cosine similarity to express the BMU selection objective:

\begin{equation}
    BMU = \argmax_{i}\:\:\cos(\theta_{x,w_i}) \label{new_rule}
\end{equation}

\noindent
where $\theta_{x,w_i}$ is the angle between the input vector $\mathbf{x}$ and the weight vector of the $i$-th unit $\mathbf{w_i}$. The cosine similarity between two vectors $\mathbf{v}$, $\mathbf{u}$ can be calculated by:

\begin{equation}
    \cos(\theta_{v,u}) = \frac{\mathbf{v}^\text{T} \mathbf{u}}{||\mathbf{v}||_2\:||\mathbf{u}||_2}
\end{equation}

\subsection{Learning Algorithm}
\begin{figure}[ht]
	\includegraphics[keepaspectratio,width=0.45\textwidth]{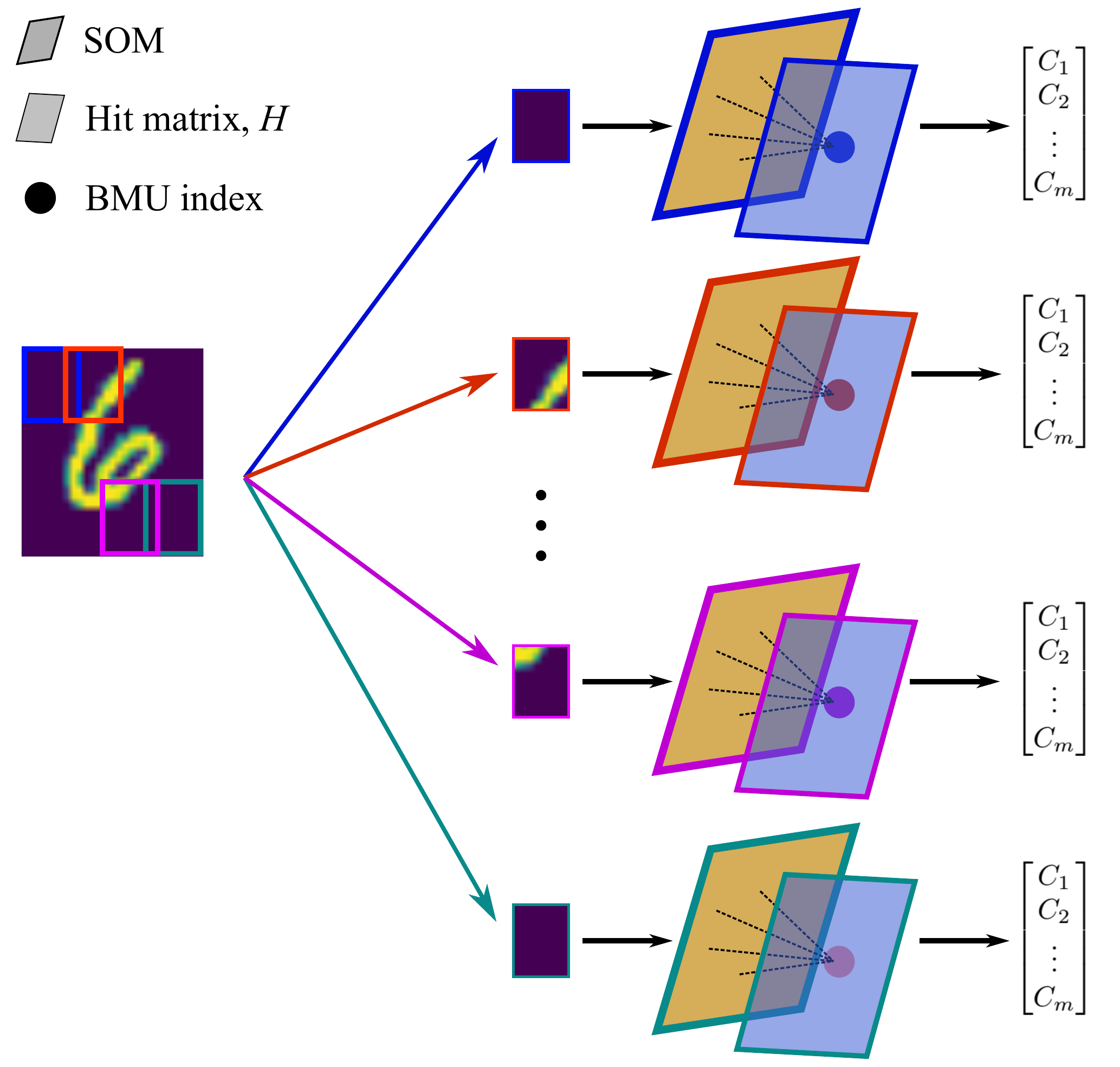}
	\caption{The input image is divided into subregions. Each subregion of the input space is modeled from the corresponding SOM. The SOMs identify the BMUs whose indices are then propagated to the appropriate hit matrices. These matrices estimate the association between the BMUs and the class labels.}
	\label{fig:dend_som}
\end{figure}

Each SOM can be trained independently by utilizing equation 7 for BMU identification, and equations 2-5 for adjusting the weight vectors. Apart from that, a hit matrix has been assigned to each SOM to assist in measuring the association between SOM units and labels. This association corresponds to the number of times a unit was selected as a BMU for a specific label (Figure \ref{fig:dend_som} and Algorithm \ref{alg:dendsom}). Notably, the learning procedure is the same for both classification and CL. However, in CL we feed the data of each class sequentially instead of sampling a random input vector from the entire dataset.

\begin{algorithm}
\caption{DendSOM training algorithms}
\label{alg:dendsom}
\textbf{Input:} $X_{train}$, $y_{train}$, $model$\\
\textbf{Parameters:} $\alpha_0$, $\sigma_0$, $\lambda$, $\alpha_{crit}$, $r_{exp}$, $n_{iter}$, $iter\_crit$\\
\textbf{Output:} $model$, $H$
\begin{algorithmic}[1]
\Statex
\State $model.random\_init()$ \Comment{Random initialization}
\State $H \gets zeros(n\_labels, S_1 \times S_2, U_1 \times U_2)$
\State $t \gets 0$
\State $iter\_crit \gets \log\big\lfloor \lambda \frac{\alpha_0}{\alpha_{crit}} \big\rfloor$
\For{$i \gets 1 $ to $n_{iter}$}
\State $sample, label \gets X_{train}[i], y_{train}[i]$
\State $rfs \gets create\_receptive\_fields(sample)$
\State $bmus \gets identify\_bmus(rfs, model)$
\State $model \gets model\_update(model, bmus, \alpha_0, \sigma_0, \lambda)$
\State $H[label, bmus] \gets H[label, bmus]+1$
\If{$i \mathbin{\%} iter\_crit == 0$}
    \State $t \gets \big\lfloor \frac{t}{r_{exp}} \big\rfloor$
\EndIf
\EndFor
\State $t=t+1$
\end{algorithmic}
\end{algorithm}

\noindent
We set $r_{exp}=1$ for classification or clustering. The $iter\_crit$ parameter can be derived as follows:
\begin{equation*}
\begin{split}
    \alpha_{crit} = \alpha_0 \text{exp}\bigg(-\frac{iter\_crit}{\lambda}\bigg)
\end{split}
\end{equation*}
\begin{equation*}
\begin{split}
    \frac{\alpha_0}{\alpha_{crit}} &= \text{exp}\bigg(\frac{iter\_crit}{\lambda}\bigg)
\end{split}
\end{equation*}
\begin{equation*}
\begin{split}
    iter\_crit &= \lambda\log\frac{\alpha_0}{\alpha_{crit}}.
\end{split}
\end{equation*}

\noindent
Applying the floor operation, we obtain:

\begin{equation}
iter\_crit = \bigg\lfloor\lambda\log\frac{\alpha_0}{\alpha_{crit}}\bigg\rfloor, iter\_crit \in \mathbb{N}
\end{equation}

\subsection{Decision Rule}
Point-wise mutual information (PMI) is a measure of the association between a pattern and an action. The PMI of two outcomes $x_o,\:y_o$ can take positive values when $x_o,\:y_o$ co-occur more frequently than expected under an independence assumption, negative values when the outcomes co-occur less frequently than expected, and it can be equal to $0$ when the outcomes are statistically independent. Given a hit matrix, we can calculate the PMI between a label $(l)$ and a BMU as follows:

\begin{equation}
    \begin{split}
    PMI(l;BMU) = \log{\frac{P(l|BMU)}{P(l)}}
    \end{split}
\end{equation}

\noindent
where $P(l|BMU)$ corresponds to the conditional probability of a class label given a BMU and it can be calculated using the hit matrix, i.e., $H$:
\begin{equation}
    \begin{split}
    P(l|BMU) = \frac{H[l,BMU]}{\sum_{i\in Labels}H[i,BMU]}
    \end{split}
\end{equation}

Moreover, the prior probability $P(l)$ can be calculated by:
\begin{equation}
    \begin{split}
    P(l) = \frac{\sum_{u\in Units}H[l,u]}{\sum_{u\in Units}\sum_{i\in Labels}H[i,u]}
    \end{split}
\end{equation}

According to Section \ref{Dend SOM}, in the case of DendSOM, we have more than one BMU. In fact, we have a BMU for each dendritic SOM, hence the problem of class prediction can be formulated in the following way. Let $I$ be an input image of unknown label and $BMU = \{bmu_1,\:bmu_2, \dots, bmu_n\}$ a set that contains the BMUs of $I$. Thus, we can express the objective of the label prediction problem as:

\begin{equation}
    \begin{split}
    prediction = \argmax_{l\in Labels}\sum_{i=1}^{n}PMI(l;bmu_i) 
    \end{split}
\end{equation}

\subsection{Addressing the challenges of CL}
The challenges that can arise during a CL task and how DendSOM addresses these challenges are described below:
\subsubsection{Catastrophic Forgetting}
As previously mentioned, catastrophic forgetting refers to a model’s inability to retain the knowledge acquired from previous concepts when trained incrementally on learning new ones. Given that DendSOM performs constrained k-means (it preserves the topological structure \cite{cottrell2016theoretical}), similar samples are matched to the same or neighboring units, while dissimilar samples are matched to different units. This strategy inherently minimizes the interference between samples and consequently reduces catastrophic forgetting.
\subsubsection{Memory Management}
Retaining knowledge is crucial for avoiding catastrophic forgetting during training in CL settings. Hence, many algorithms utilize memory mechanisms to store and replay information about past tasks \cite{rolnick2018experience,chaudhry2019continual}. An efficient memory management system should only save valuable information that can be transferred to future tasks. The trade-off problem, also known as the plasticity-stability \cite{mermillod2013stability} problem, states that an algorithm should balance the information saved and forgotten. The DendSOM is a memoryless architecture with respect to example data, and thus, memory management has no impact on the performance of our algorithm. However, minor memory usage is required for storing the hit matrices and the DendSOM's units. 
\subsubsection{Concept Drift}
Distribution Shifts are caused by distribution changes that may lead to catastrophic interference and consequently to catastrophic forgetting. Hence, a model has to detect and adapt to these changes to alleviate the effects of distributional shifts. There are two types of concept drifts \cite{barros2018large}. Changes in the input distribution cause the virtual concept drift, and the actual concept drift is prompted by novelty in the data. The DendSOM model accounts for both real and virtual concept drifts. It does so by estimating the prior and posterior probabilities and utilizing PMI as a decision rule since it is only affected by the ratio of the posterior and prior probabilities and not by their exact values.

\section{Benchmark Datasets \& The Split Protocol}
We evaluated our network on the MNIST, the MNIST-FASHION, and the CIFAR-10 datasets, all of which are used to benchmark CL problems.

\begin{figure}[!ht]
	\centering
	\includegraphics[keepaspectratio,width=0.5\textwidth]{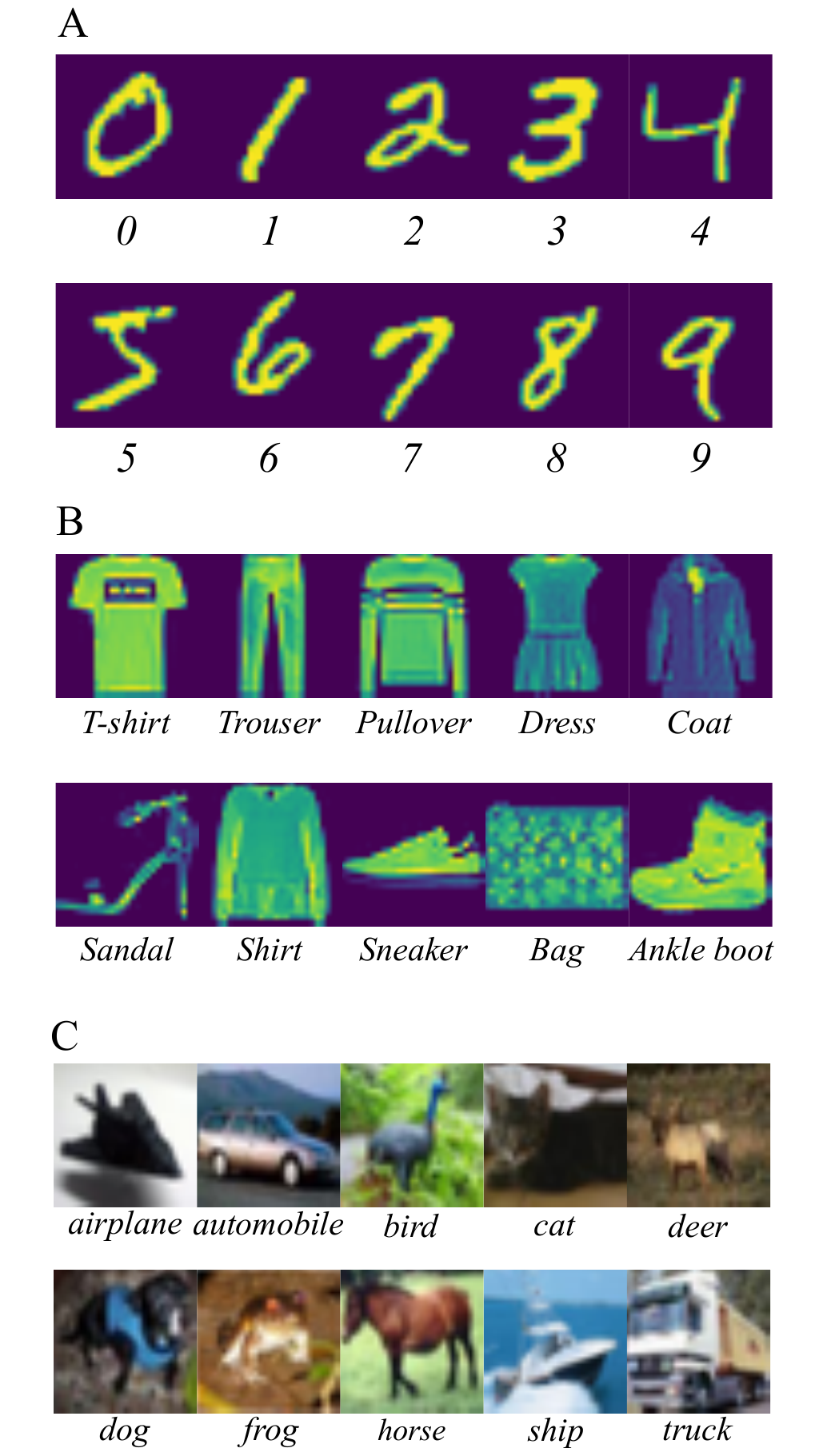}
	\caption{Example images for the different classes of the \textbf{A.}  MNIST dataset, \textbf{B.}  MNIST-FASHION dataset and \textbf{C.} CIFAR-10 dataset.}
	\label{fig:datasets}
\end{figure}

\subsection{MNIST}
The MNIST dataset \cite{lecun1998mnist} contains images of handwritten digits and consists of a training set with 60,000 samples and a test set with 10,000 samples. Each image is centered and size-normalized to $28\times 28$ pixels with 256 gray levels and is associated with a label from 10 classes (i.e., 0-9 digits) (Figure \ref{fig:datasets}A). 
\subsection{MNIST-FASHION}
The MNIST-FASHION dataset \cite{xiao2017fashion} contains images of clothes and serves as a replacement of the original MNIST dataset. In particular, these datasets share the same image size, labeling, and structure of training and testing splits (Figure \ref{fig:datasets}B).
\subsection{CIFAR-10}
The CIFAR-10 dataset \cite{krizhevsky2009learning} consists of 60,000 $32 \times 32$ color images in 10 classes, with 6,000 images per class. The training set consists of 50,000 samples, while there are 10,000 test images. In our implementation, each image is turned into gray-scale since our model can not yet process color images (i.e., three-dimensional images) (Figure \ref{fig:datasets}C).

\subsection{The Split protocol}
\begin{figure}[!ht]
	\centering
	\includegraphics[keepaspectratio,width=0.5\textwidth]{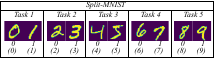}
	\caption{Tasks of the Split-MNIST protocol. The digits in parenthesis below each image denote the original label, while the digits not in parenthesis denote the label with respect to a given task, i.e., binary classification.}
	\label{fig:split_mnist}
\end{figure}

The split protocol operates on the entire dataset by splitting the given dataset into clearly separated tasks (Figure \ref{fig:split_mnist}). The Split MNIST is a typical dataset that is utilized for benchmarking CL methods. It was introduced in a five-task form where the ten classes are split into five binary classification problems. We also apply the split protocol to the CIFAR-10 dataset, which is significantly more challenging.

\section{CL scenarios}
Although several machine learning scenarios permit models to operate on the entire dataset, in CL, data arrive incrementally as subsets of samples. According to \cite{van2019three}, CL cases are categorized into three main scenarios. Using the Split-MNIST as an example, we summarize the objective of each scenario in Table \ref{tab:cl_scenarios}: 

\begin{table}[!ht]
    \caption{Objective of each CL scenario. The scenarios are shown from the easiest to the most difficult one.}
    \label{tab:cl_scenarios}
    \centering
    \begin{tabular*}{1.0\linewidth}{ll}
        \Xhline{1.0pt}
        \bfseries Scenario & \bfseries Objective\\
        \Xhline{0.5pt}
        Task-IL & Is it 0 or 1? (Task given)\\
        Domain-IL & Is it 0 or 1? (Task unknown)\\
        Class-IL & Which digit is it? (0-9)\\
        \Xhline{1.0pt}
    \end{tabular*}
\end{table}
 
\subsection{Task-Incremental Learning}
\begin{figure}[!ht]
	\centering
	\includegraphics[keepaspectratio,width=0.5\textwidth]{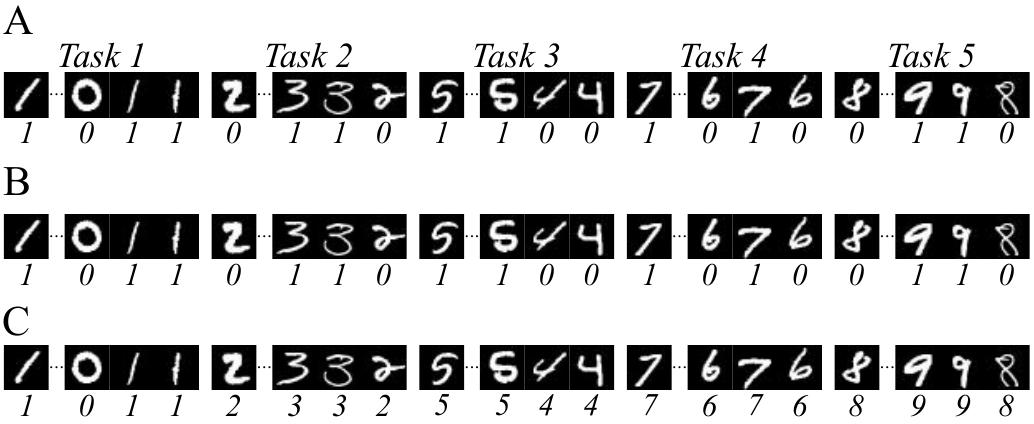}
	\caption{Illustration of the different CL scenarios using the split-MNIST dataset. \textbf{A.} Task-IL scenario. \textbf{B.} Domain-IL scenario. \textbf{C.} Class-IL scenario. At bottom of each image the target class is given. Notice that in \textbf{B} and \textbf{C} the task is not given to the model.}
	\label{fig:scenarios_cl_mnist}
\end{figure}

Task-Incremental Learning (Task-IL) is the easiest CL scenario since the model can utilize task-related information. The dataset is permuted into non-overlapping tasks, and during training, only data from the current task is available (Figure \ref{fig:scenarios_cl_mnist}A).

\subsection{Domain-Incremental Learning}
Domain-Incremental Learning (Domain-IL) is similar to Task-IL except that the model cannot access information related to the task. Thus, for a given set $D=\{x_i,y_i\}_{i=1...2}$, the model has to solve the task at hand (Figure \ref{fig:scenarios_cl_mnist}B).

\subsection{Class-Incremental Learning}
\begin{figure}[!ht]
	\centering
	\includegraphics[keepaspectratio,width=0.5\textwidth]{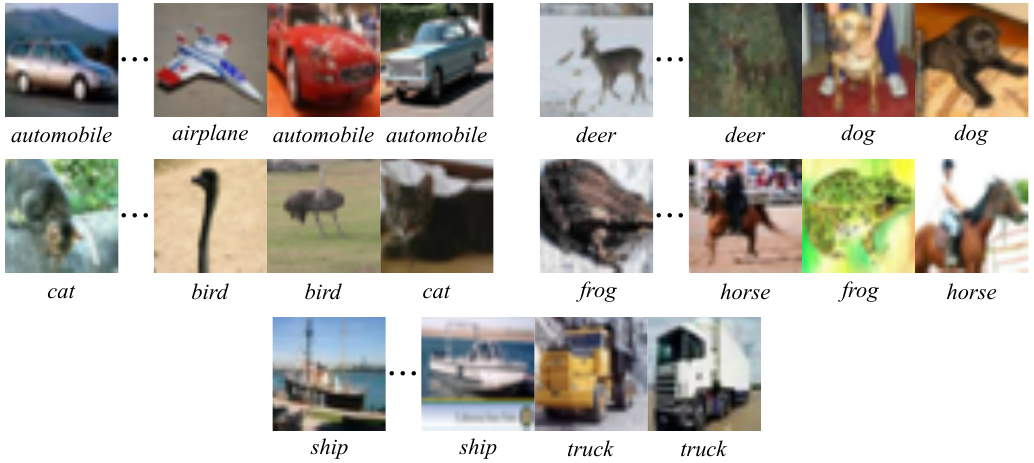}
	\caption{Class-IL scenario on the Split-CIFAR-10 dataset. On bottom of each image the target class is given.}
	\label{fig:scenarios_cl_cifar10}
\end{figure}

The goal of Class-Incremental Learning (Class-IL) is to learn, given a dataset $D=\{x_i,y_i\}_{i=1...n}$, a unified classifier. Similar to Domain-IL, the model can not utilize task information in this scenario (Figure \ref{fig:scenarios_cl_mnist}C and \ref{fig:scenarios_cl_cifar10}).

\section{Parameter Settings}
To diminish the non-deterministic effect of the initialization phase (random initial weights), we repeated each experiment 10 times (i.e., trials). This number of trials is sufficient to illustrate the behavior of our proposed algorithm. We selected the architecture and training parameters for the network models through a trial-and-error process. It is worth noting that $U_1$ and $U_2$ correspond to the number of units in each SOM. For comparison purposes, we report the performance of DendSOM against the one achieved by a traditional SOM network with a similar number of trainable parameters.
\subsection{Training Parameters}
The DendSOM training hyperparameters for both SOM vs. DendSOM comparison and incremental-learning experiments are depicted in Table \ref{tab:tr_hpr}. Note that only the hyperparameter $r_{exp}$ changes in the case of incremental learning experiments since it prevents the algorithm from converging before the completion of the training process.

\begin{table}[!ht]
    \caption{Training hyperparameters for the classification task, i.e., SOM vs. DendSOM.}
    \label{tab:tr_hpr}
    \centering
    \begin{tabular*}{1.0\linewidth}{l|cccc}
        \Xhline{1.0pt}
        \bf{Dataset} & $\mathbf{\alpha_0}$ & $\mathbf{\sigma_0}$ & $\mathbf{\lambda}$ & $\mathbf{r_{exp}}$\\
        \Xhline{0.5pt}
    	MNIST & $0.95$ & $4$ & $1\text{e}3$ & $1^{\dagger}$, $2^{\ddagger}$\\
    	MNIST-FASHION & $0.95$ & $5$ & $1\text{e}3$ & $1^{\dagger}$, $2^{\ddagger}$\\
    	CIFAR-10 & $0.95$ & $6$ & $1\text{e}3$ & $1^{\dagger}$, $2^{\ddagger}$\\
        \Xhline{1.0pt}
    	\multicolumn{5}{l}{$^{\dagger}$ classification task.}\\
    	\multicolumn{5}{l}{$^{\ddagger}$ incremental learning task.}
    \end{tabular*}
\end{table}

\subsection{Architecture Parameters}
The architecture parameters used on the MNIST, MNIST-FASHION, and CIFAR-10 datasets are shown in Table \ref{tab:model_hpr}. We used the same DendSOM architecture in all of the reported experiments.

\begin{table}[!ht]
    \caption{Architecture parameters for each architecture and dataset.}
    \label{tab:model_hpr}
    \centering
    \begin{tabular*}{1.0\linewidth}{lcccccccc}
        \Xhline{1.0pt}
        \multicolumn{9}{l}{\bf{MNIST}}\\
        \Xhline{1.0pt}
    	\textbf{Model} & $\mathbf{S_1}$ & $\mathbf{S_2}$ &$\mathbf{U_1}$ & $\mathbf{U_2}$ &$\mathbf{P_1}$ & $\mathbf{P_2}$ & $\mathbf{s_1}$ & $\mathbf{s_2}$\\
        \Xhline{0.5pt}
    	DendSOM & $7$ & $7$ &$8$ & $8$ &$10$ & $10$ & $3$ & $3$\\
    	SOM & $1$ & $1$ &$21$ & $21$ &$28$ & $28$ & $-$ & $-$\\
    	\Xhline{1.0pt}
    	\\
    	\Xhline{1.0pt}
        \multicolumn{9}{l}{\bf{MNIST-FASHION}}\\
        \Xhline{1.0pt}
    	\textbf{Model} & $\mathbf{S_1}$ & $\mathbf{S_2}$ &$\mathbf{U_1}$ & $\mathbf{U_2}$ &$\mathbf{P_1}$ & $\mathbf{P_2}$ & $\mathbf{s_1}$ & $\mathbf{s_2}$\\
        \Xhline{0.5pt}
    	DendSOM & $6$ & $6$ &$10$ & $10$ &$8$ & $8$ & $4$ & $4$\\
    	SOM & $1$ & $1$ &$18$ & $18$ &$28$ & $28$ & $-$ & $-$\\
    	\Xhline{1.0pt}
    	\\
    	\Xhline{1.0pt}
        \multicolumn{9}{l}{\bf{CIFAR-10}}\\
        \Xhline{1.0pt}
    	\textbf{Model} & $\mathbf{S_1}$ & $\mathbf{S_2}$ &$\mathbf{U_1}$ & $\mathbf{U_2}$ &$\mathbf{P_1}$ & $\mathbf{P_2}$ & $\mathbf{s_1}$ & $\mathbf{s_2}$\\
        \Xhline{0.5pt}
    	DendSOM & $15$ & $15$ &$12$ & $12$ &$4$ & $4$ & $2$ & $2$\\
    	SOM & $1$ & $1$ &$29$ & $29$ &$32$ & $32$ & $-$ & $-$\\
    	\Xhline{1.0pt}
    \end{tabular*}
\end{table}

\subsection{Computational resources and Statistical tests}
The experiments presented in the following section and the respective analysis are implemented in python (version 3.8.8) programming language using custom-based software. The scripts and derived data supporting the findings of this study are available from the corresponding author [P.P] on request. All experiments were performed on a desktop equipped with an Intel${\circledR}$ Core$^{\text{TM}}$ i7-9700 CPU, a GeForce RTX 2080 Ti GPU, 32 GB RAM, and CentOS Linux (version 7.9) operating system. To compare the algorithms, we performed Welch's ANOVA test. The $\alpha$ was set to $0.05$ ($95\%$ confidence interval). To account for multiple comparisons, we performed a Dunnett's T3 test.

\section{Experiments}
\subsection{SOM vs. DendSOM}
\subsubsection{Experiment Objective}
In this experiment, we test the hypothesis that the DendSOM algorithm allows for improved classification performance compared to the original SOM algorithm. To do so, we compare the two algorithms in terms of the accuracy score on three publicly available datasets.

\begin{table}[!ht]
    \caption{Classification accuracy scores obtained by each architecture on three benchmark datasets. Performance accuracy is listed as mean $\pm$ standard deviation over $N=10$ trials. All comparisons amongst all algorithms are statistically significant ($p<10^{-3}$, Dunnett's T3 multiple comparisons test) for MNIST, MNIST-FASHION, and CIFAR-10 datasets, respectively.}
    \label{tab:unsclass}
    \centering
    \begin{tabular*}{1.0\linewidth}{lccc}
        \Xhline{1.0pt}
        \textbf{Model} & \textbf{MNIST} & \textbf{MNIST-FASHION} & \textbf{CIFAR-10}\\
        \Xhline{0.5pt}
        SOM$^\dagger$ & $84.94\pm 0.58$& $73.94\pm 0.34$ & $26.45\pm 0.13$\\
        SOM$^\ddagger$ & $87.26\pm 1.07$ & $76.09\pm 0.29$ & $27.13 \pm 0.15$\\
        \textbf{DendSOM}$^\dagger$ & $\mathbf{94.93\pm 0.18}$ & $\mathbf{77.84\pm0.11}$ & $\mathbf{37.97 \pm 0.11}$\\
        \textbf{DendSOM}$^\ddagger$ & $\mathbf{95.28\pm 0.02}$ & $\mathbf{80.88\pm 0.26}$ & $\mathbf{46.79\pm0.10}$\\
        \Xhline{1.0pt}
        \multicolumn{4}{l}{$^\dagger$ Euclidean distance was used for BMU identification.}\\
        \multicolumn{4}{l}{$^\ddagger$ Cosine similarity was used for BMU identification.}
    \end{tabular*}
\end{table}

\subsubsection{Results}
As evident from the results in Table \ref{tab:unsclass}, both the architecture and the distance used for BMU identification affect performance. Specifically, utilizing multiple SOMs to map different regions of the image (DendSOM) improves the performance of the original algorithm (SOM) since the final matching pattern is not just a weighted average of similar images but consists of several informative features. Furthermore, using the cosine similarity as opposed to the Euclidean distance for the identification of the BMU allows for improved performance as the former is only influenced by the inter-pixel relation between the weight vector and the corresponding receptive field. In contrast, the Euclidean distance takes into account the average difference between pixel values. Specifically, the DendSOM that employs cosine-similarity for BMU identification achieves the highest accuracy score on all datasets. Finally, all algorithms achieve their highest and lowest scores on the MNIST and CIFAR-10 datasets, respectively, which is expected because the MNIST dataset consists of handwritten digits in a static background (and therefore is the simplest task), while the CIFAR-10 dataset consists of 10 real-world classes in a dynamic background. Due to this complexity, the geometrical properties of the input space cannot be described appropriately neither by the cosine-similarity nor by the Euclidean distance.

\subsection{Task-IL scenario}
\subsubsection{Experiment Objective}
The performance of our algorithm is also evaluated and compared to other state-of-the-art architectures \cite{van2019three} on the Split-MNIST protocol under the Task-IL scenario. The results are shown in Table \ref{tab:tilr}, and the comparison of the different models in Figure \ref{fig:performance_cl}A.

\begin{table}[!ht]
    \caption{Accuracy scores for all compared algorithms on the Split-MNIST protocol under the Task-IL scenario. Performance accuracy is listed as mean $\pm$ standard deviation over $N=10$ (SOM, DendSOM) and $N=20$ (for the other algorithms) trials. All comparisons between DendSOM and the other algorithms are statistically significant ($p<10^{-4}$, Dunnett's T3 multiple comparisons test).}
    \label{tab:tilr}
    \centering
    \begin{tabular*}{1.0\linewidth}{lc}
        \Xhline{1.0pt}
        \textbf{Model} & \textbf{MNIST}\\
        \Xhline{0.5pt}
		SOM$^\dagger$ & $91.87\pm 0.58$\\
		SOM$^\ddagger$ & $92.82\pm 0.36$\\
		\textbf{DendSOM}$^\ddagger$ & $\mathbf{97.73\pm 0.25}$\\
		EWC & $98.64\pm 0.22$\\
		SI & $99.09\pm 0.15$\\
		XdG & $99.10\pm 0.08$\\
		Online EWC & $99.12\pm 0.11$\\
		DGR & $99.50\pm 0.03$\\
		LwF & $99.57\pm 0.02$\\
		DGR+ distill & $99.61\pm 0.02$\\
        \Xhline{1.0pt}
        \multicolumn{2}{l}{$^\dagger$ Euclidean distance was used for BMU identification.} \\
        \multicolumn{2}{l}{$^\ddagger$ Cosine similarity was used for BMU identification.}
    \end{tabular*}
\end{table}

\subsubsection{Results}
As shown in Table \ref{tab:tilr}, most models perform well in this scenario, except for the SOM algorithm. Notably, the DendSOM algorithm achieves a higher accuracy score than the SOM algorithm because the latter process the whole image at once. Importantly, unlike other methods, neither the SOM nor the DendSOM can use the task information to learn task-specific features by freezing or regularizing their weight vectors. In contrast, they utilize this information only to estimate the appropriate probability distributions. In addition, when the Euclidean distance is used as a BMU selection rule, the algorithm does not consider the temporal evolution of the images.

\subsection{Domain-IL scenario}
\subsubsection{Experiment Objective}
This experiment is similar to the one mentioned above except that the models operate under Domain-IL constraints \cite{van2019three}.

\subsubsection{Results}
\noindent
As shown in Table \ref{tab:dilr} and in Figure \ref{fig:performance_cl}B, the $DendSOM_{cos}$ algorithm achieves an accuracy score of $89.70$, ranking among the top 3 models for Domain-IL on the Split-MNIST dataset. However, the other algorithms perform multiple training epochs per task and are trained in batch mode. On the contrary, there is only one stream of samples, and each sample is presented only once in both the SOM and DendSOM algorithms. While the SOM algorithm using the Euclidean distance ranks fifth with respect to performance accuracy, its performance is inferior to the DendSOM algorithm with cosine similarity. The superiority of the DendSOM algorithm can be explained by its architecture and the BMU selection rule, as mentioned in the previous experiments.

\begin{table}[!ht]
    \caption{Accuracy scores for all compared algorithms on the Split-MNIST protocol under the Domain-IL scenario. Performance accuracy is listed as mean $\pm$ standard deviation over $N=10$ (SOM, DendSOM) and $N=20$ (for the other algorithms) repetitions. All comparisons between DendSOM and the other algorithms are statistically significant ($p<10^{-4}$, Dunnett's T3 multiple comparisons test).}
    \label{tab:dilr}
    \centering
    \begin{tabular*}{1.0\linewidth}{lc}
        \Xhline{1.0pt}
        \textbf{Model} & \textbf{MNIST}\\
        \Xhline{0.5pt}
		SOM$^\dagger$ & $78.69\pm 1.28$\\
		SOM$^\ddagger$ & $81.90\pm 1.04$\\
		\textbf{DendSOM}$^\ddagger$ & $\mathbf{89.70\pm 0.27}$\\
		EWC & $63.95\pm 1.90$\\
		Online EWC & $64.32\pm 1.90$\\
		SI & $65.36\pm 1.57$\\
		LwF & $71.50\pm 1.63$\\
		DGR & $95.72\pm 0.25$\\
		DGR+ distill & $96.83\pm 0.20$\\
        \Xhline{1.0pt}
        \multicolumn{2}{l}{$^\dagger$ Euclidean distance was used for BMU identification.}\\
        \multicolumn{2}{l}{$^\ddagger$ Cosine similarity was used for BMU identification.}
    \end{tabular*}
\end{table}

\subsection{Class-IL scenario}

\subsubsection{Experiment Objective}
\begin{figure}[!ht]
	\centering
	\includegraphics[keepaspectratio,width=0.5\textwidth]{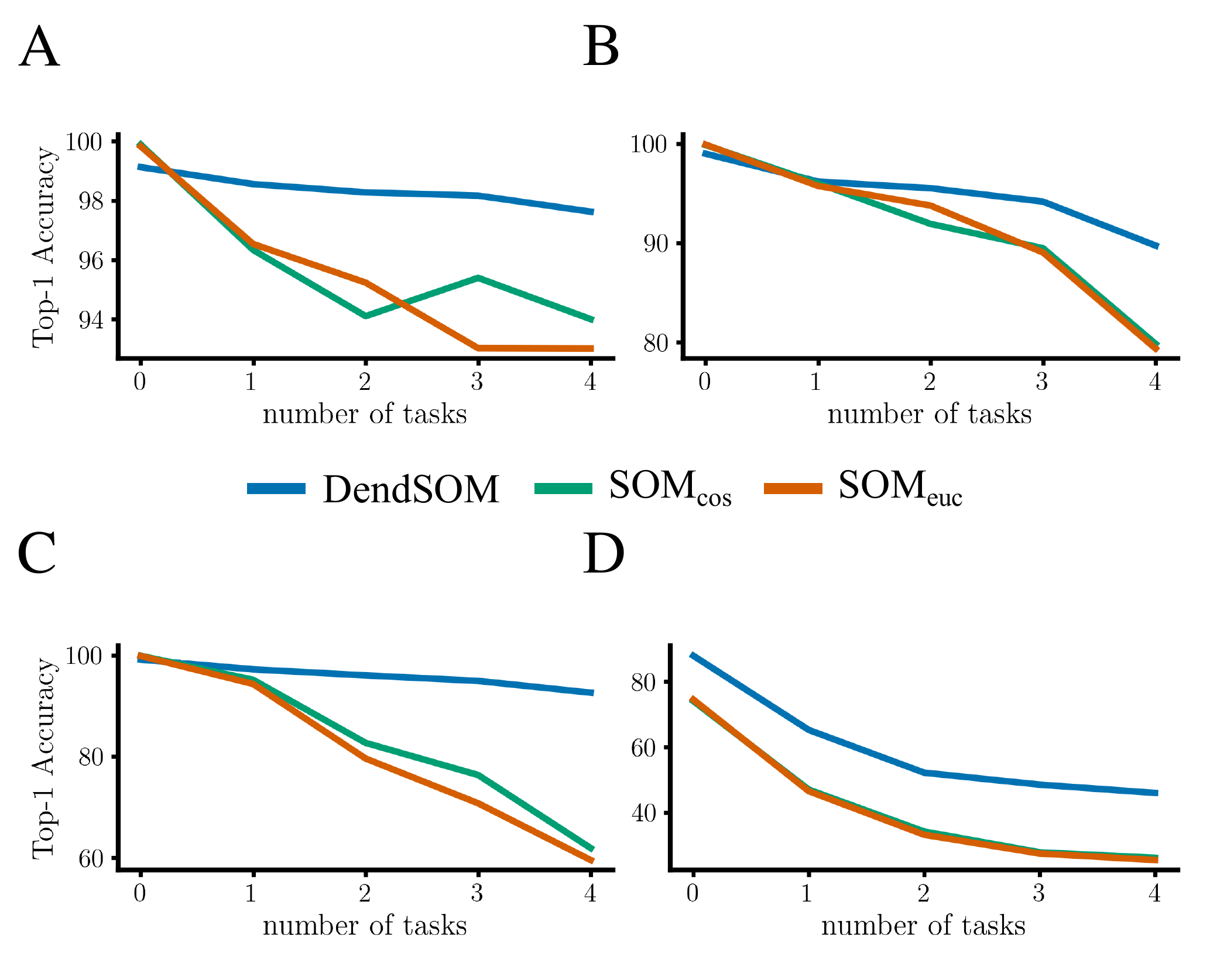}
	\caption{Illustration of the model performance for different CL scenarios using the split-MNIST and Split-CIFAR-10 datasets over 3 repetitions per scenario and model. y-axis shows the Top-1 Accuracy, i.e.,  the conventional accuracy, while x-axis denote the number of tasks. (blue) DendSOM, (orange) SOM with Euclidean distance for BMU identification, (green) SOM with cosine similarity for BMU identification. \textbf{A.} Task-IL scenario. \textbf{B.} Domain-IL scenario. \textbf{C.} Class-IL scenario using the Split-MNIST. \textbf{D.} Class-IL scenario using the Split-CIFAR-10.}
	\label{fig:performance_cl}
\end{figure}

Finally, we compare the performance of our algorithm to the state-of-the-art models \cite{van2021class} in Class-IL scenario, on both Split-MNIST (Figure \ref{fig:performance_cl}C) and Split-CIFAR-10 (Figure \ref{fig:performance_cl}D) datasets.

\subsubsection{Results}
As shown in Table \ref{tab:cilr}, DendSOM with cosine similarity outperforms most of the other algorithms for both datasets. The DendSOM algorithm achieved accuracy scores of $92.82$ and $45.97$ on the Split-MNIST and Split-CIFAR-10 datasets, respectively, corresponding to the second-highest scoring algorithm. Furthermore, unlike the other models, the DendSOM with cosine similarity was trained online since the tasks were presented sequentially and only once, while the batch size was equal to 1. Once again, the SOM-based algorithms are proven to be simple yet effective algorithms in CL that perform comparably with several leading-edge regularization and hybrid algorithms. However, the DendSOM with cosine similarity achieves a significantly higher accuracy score than the SOM with euclidean distance because the cosine similarity rule does not consider the sum of individual pixel intensity differences is the case when euclidean distance is used. Moreover, the restrictive connectivity introduced in the DendSOM algorithm with cosine similarity allows for a more detailed mapping of the input space, which, in turn, accounts for capturing the temporal changes in the input data (images).

\begin{table}[!ht]
    \caption{Accuracy scores for all compared algorithms on the Split-MNIST and Split-CIFAR-10 protocols under the Class-IL scenario. Performance accuracy is listed as mean $\pm$ standard deviation over $N=10$ trials. All comparisons between DendSOM and the other algorithms are statistically significant ($p<10^{-4}$, Dunnett's T3 multiple comparisons test) for both Split-MNIST and Split-CIFAR-10 protocols.}
    \label{tab:cilr}
    \centering
    \begin{tabular*}{1.0\linewidth}{lcc}
        \Xhline{1.0pt}
        \textbf{Model} & \textbf{MNIST} & \textbf{CIFAR-10}\\
        \Xhline{0.5pt}
		SOM$^\dagger$ & $59.61\pm 0.67$ & $25.55\pm 0.36$\\	
		SOM$^\ddagger$ & $62.92\pm 0.85$ & $26.43\pm 0.24$\\	
        \textbf{DendSOM}$^\ddagger$ & $\mathbf{92.82\pm 0.12}$ & $\mathbf{45.97\pm 0.07}$\\
		DGR & $91.30\pm 0.60$ & $17.21\pm 1.88$\\		
		EWC & $19.95\pm 0.05$ & $18.63\pm 0.29$\\		
		SI & $19.95\pm0.11$ & $18.14\pm 0.36$\\		
		CWR & $32.48\pm 2.64$ & $18.37\pm 1.61$\\		
		CWR+ &  $ 37.21\pm 3.11$ & $ 22.32\pm 1.08$\\	
		AR1 & $48.84\pm 2.55$ & $24.44\pm 2.55$\\		
		GC & $93.79\pm 0.08$ & $56.03 \pm 0.04$\\	
        \Xhline{1.0pt}
        \multicolumn{3}{l}{$^\dagger$ Euclidean distance was used for BMU identification.}\\
        \multicolumn{3}{l}{$^\ddagger$ Cosine similarity was used for BMU identification.}
    \end{tabular*}
\end{table}

\section{Conclusion}
In this work, we proposed a new SOM-based architecture (DendSOM) that utilizes neuroscience and information theory concepts to address the long-standing problem of catastrophic-forgetting, which is crucial for avoiding model re-training when new data are presented. We showed that DendSOM consistently outperforms the original SOM algorithm on several benchmark classification tasks. Thus, mimicking the dendritic processing observed in biological neurons (i.e., receptive fields and local computation of BMU) plays a significant role in improving the predictive power of classical SOM architectures. Importantly, DendSOM is a memory-efficient algorithm, trained for just a single epoch, using only one input stream -equivalent to a batch size of one- which further improves implementation efficiency compared to other algorithms. Finally, we showed that DendSOM performs on par with the current state-of-the-art methods on many continual learning tasks while achieving higher performance than most leading-edge architectures on the Domain-IL and Class-IL scenarios. Overall, DendSOM has several advantages in its architecture and performance accuracy, which can help address the problem of catastrophic forgetting for real-world applications.

\section*{Author contributions}
K.P. designed, implemented and performed the experiments and the data analysis. K.P., S.C. and P.P. conceived and wrote the paper. P.P. supervised the project.

\section*{Conflict of interest}
The authors declare no conflict of interest.

\section*{Acknowledgment}
We thank Grigorios Tsagkatakis, PhD for his critical and constructive feedback on our manuscript. This work was supported by the H2020-FETOPEN-2018-2019-2020-01, FET-Open Challenging Current Thinking, grant NEUREKA: GA-863245 and the NIH grant No 1R01MH124867-01.

\appendix

\section*{Hyperparameter-tuning Analysis}
Several training hyperparameters control the behavior of the DendSOM. The ultimate goal is to find an optimal combination of these hyperparameters that increases the predictive power of our model and thus gives better results. Hence, additional experiments have been performed to determine how the training hyperparameters affect the performance of our model individually. In fact, we discovered a good set of hyperparameters through a trial-and-error process (Tables \ref{tab:tr_hpr} and \ref{tab:model_hpr}) based on which we carried out the following experiments.

\begin{figure}[!ht]
	\centering
	\includegraphics[keepaspectratio,width=0.5\textwidth]{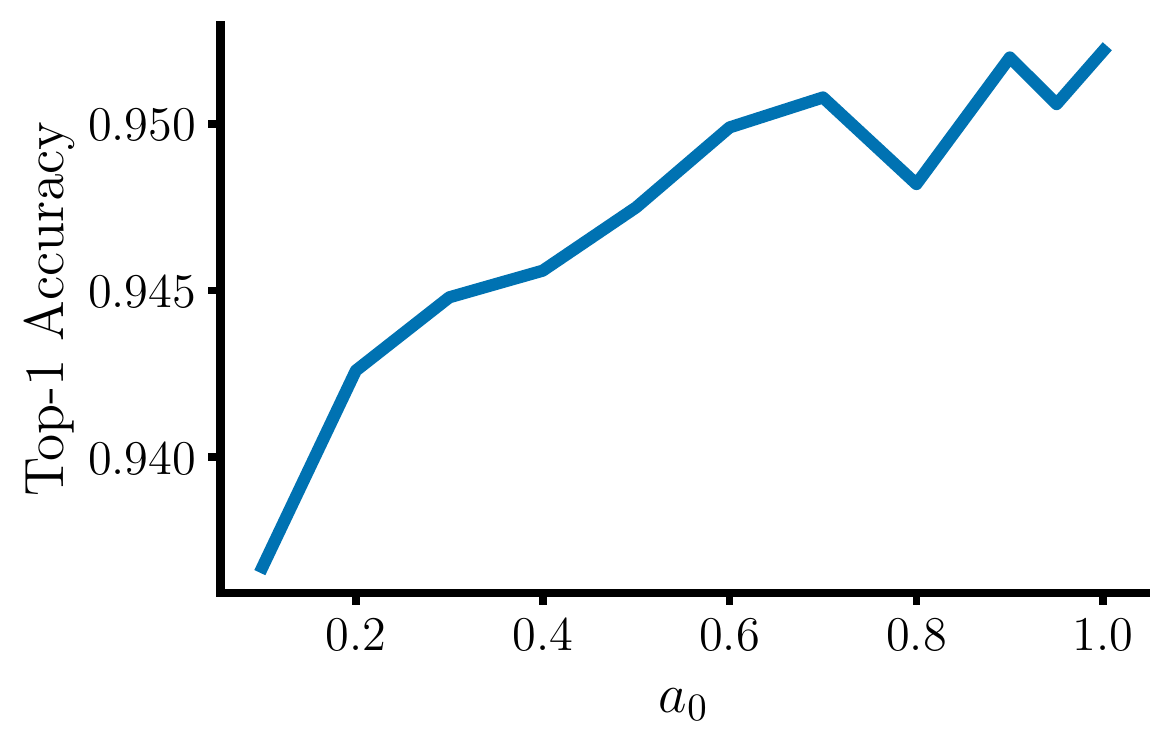}
	\caption{Initial learning rate affects the classification accuracy of the DendSOM algorithm on the MNIST dataset. As the initial learning rate ($\alpha_0$) increases, the accuracy of the model is improved.}
	\label{fig:a0_uns}
\end{figure}

As Figure \ref{fig:a0_uns} shows, a higher $\alpha_0$ usually results in a higher accuracy score. However, the original SOM algorithm requires that $0<\alpha_0<1$, and the maximum cosine similarity rule for BMU identification is a scale-invariant rule. Thus, this experiment can be extended for $\alpha_0>1$. For $\alpha_0=0.1$, the DendSOM model has an accuracy score of $0.936$, which suggests that it is possible to make accurate predictions without updating the map during the training process, but this hypothesis has to be tested.

We initialized the $\sigma_0$ using the following heuristic:
$$\sigma_0=\frac{\max\{U_1,U_2\}}{2}$$
\noindent
where $U_1$ and $U_2$ correspond to the number of units across the horizontal and vertical axis respectively

\begin{figure}[!ht]
	\centering
	\includegraphics[keepaspectratio,width=0.5\textwidth]{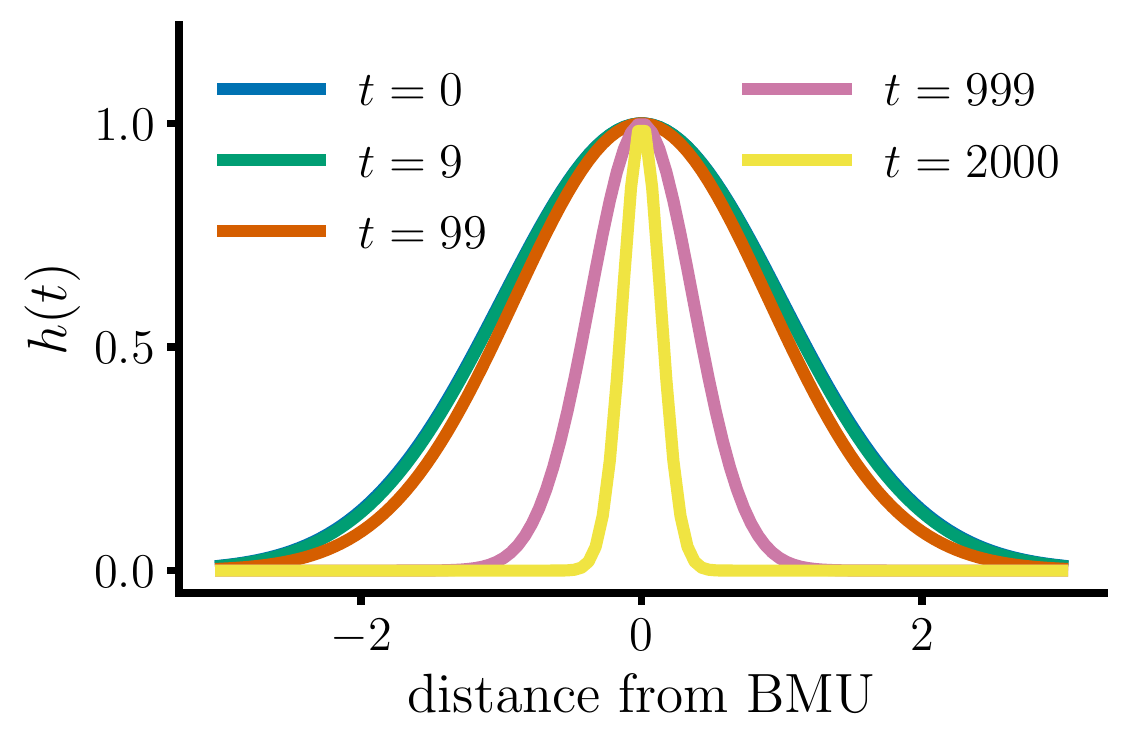}
	\caption{The behavior of $h$ functions for $r_{exp}=1$ as function of the distance from the BMU. Colors denote different values of $t$.}
	\label{fig:nf}
\end{figure}

\begin{figure}[!ht]
    \centering
    \begin{subfigure}[b]{\textwidth}
        \includegraphics[keepaspectratio,width=0.5\textwidth]{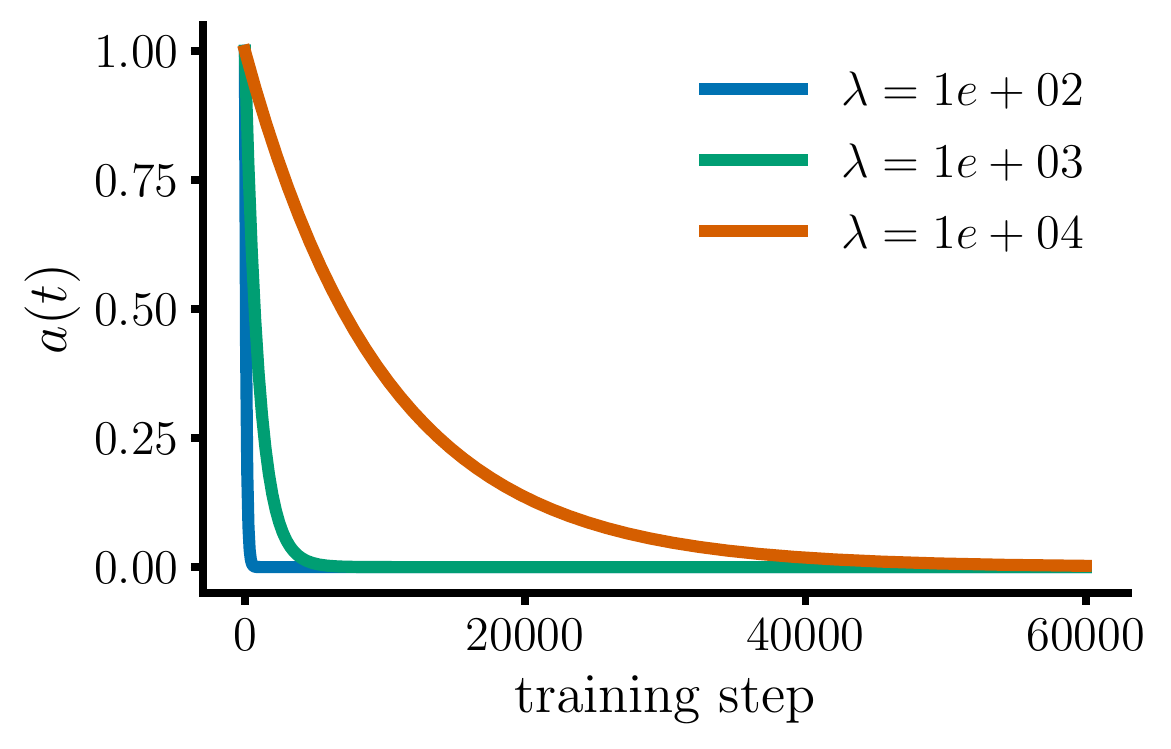}
    \end{subfigure}
    \hfill
    \begin{subfigure}[b]{\textwidth}
        \includegraphics[keepaspectratio,width=0.5\textwidth]{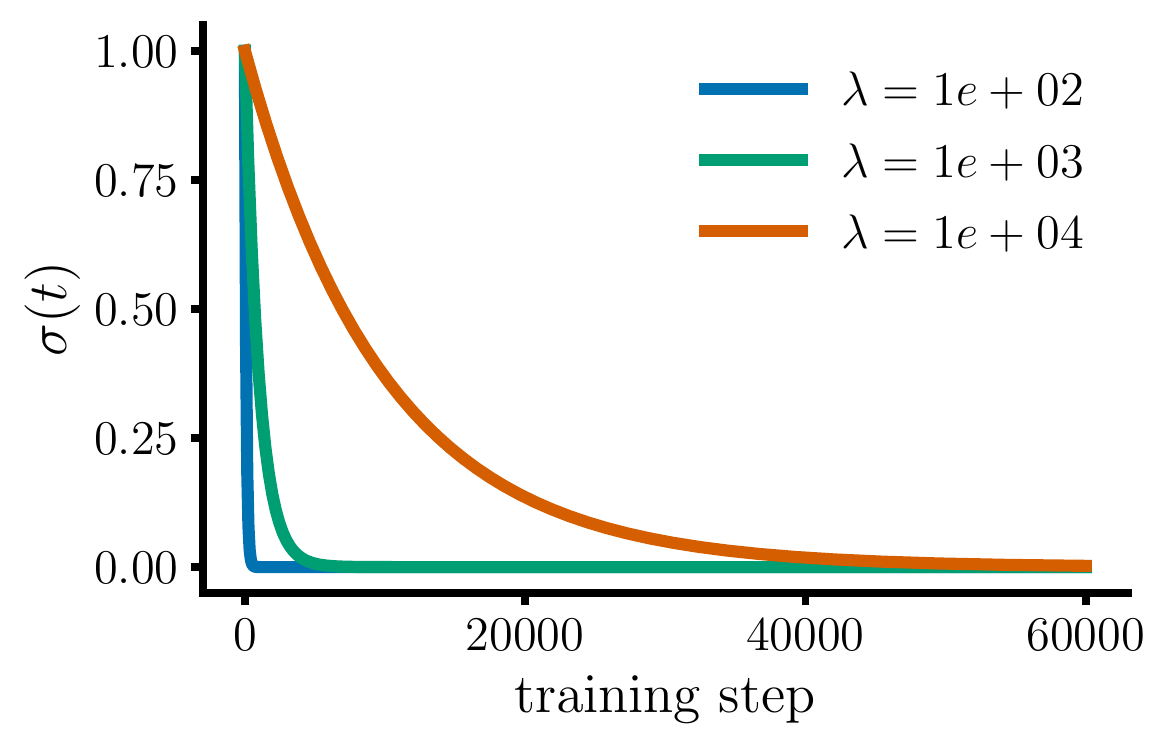}
    \end{subfigure}
    \caption{Behavior of $a$ (\textit{top}) and $\sigma$ (\textit{bottom}) as functions of the training steps. Colors denote different values of the $\lambda$ parameter.}
    \label{fig:lr_nr_decay}
\end{figure}

\begin{figure}[!ht]
    \centering
    \begin{subfigure}[b]{\textwidth}
        \includegraphics[keepaspectratio,width=0.5\textwidth]{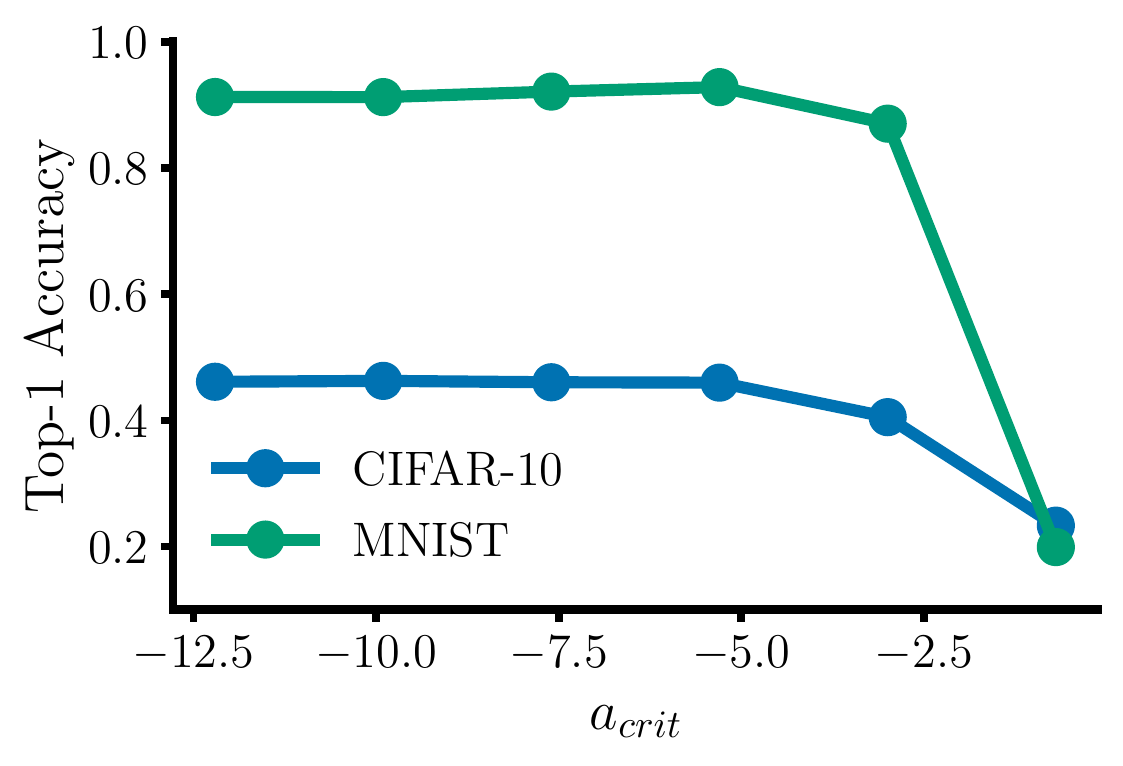}
    \end{subfigure}
    \hfill
    \begin{subfigure}[b]{\textwidth}
        \includegraphics[keepaspectratio,width=0.5\textwidth]{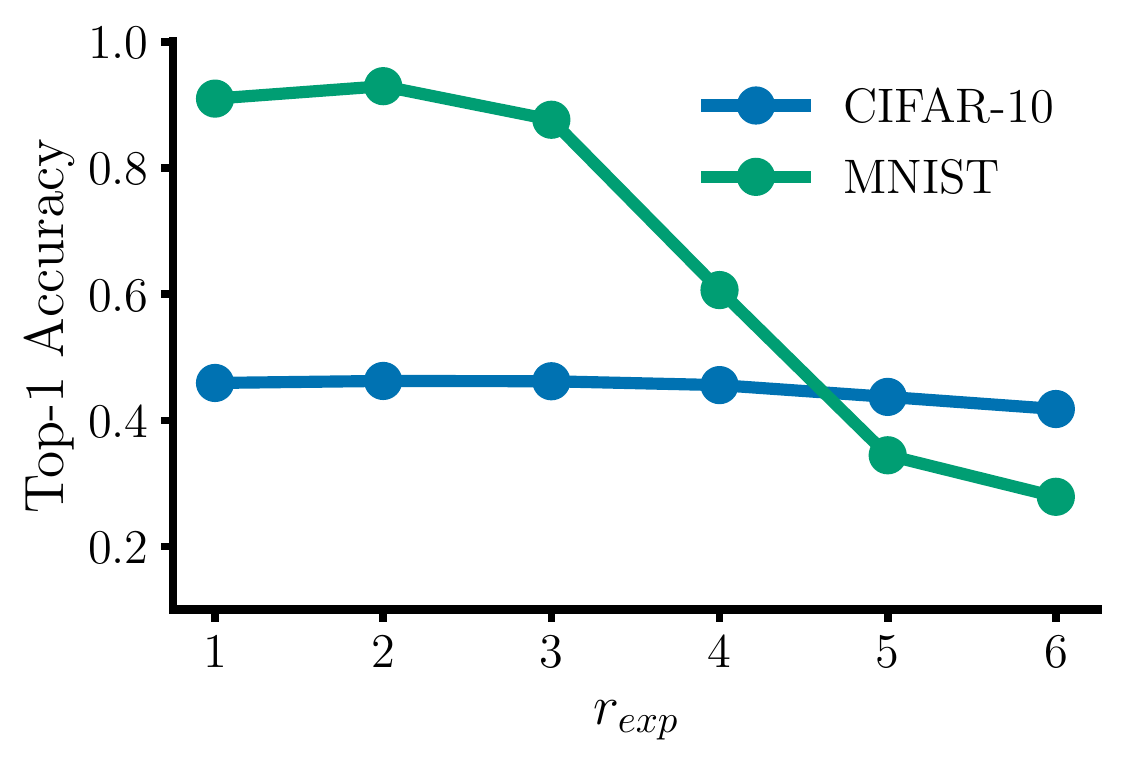}
    \end{subfigure}
    \caption{\textit{(top)} Influence of $\alpha_{crit}$ and \textit{(bottom)} $r_{exp}$ hyperparammeters over the DendSOM algorithm tested on CIFAR-10 (blue filled circles) and MNIST (green filled circles). Notice that the x-axis in the top panel is in logarithmic scale since $\alpha_{crit}\in\{5\text{e-}1,5\text{e-}2,5\text{e-}3,5\text{e-}4,5\text{e-}5,5\text{e-}6\}$.}
    \label{fig:ac_rx_cont}
\end{figure}

\begin{figure}[!ht]
	\centering
	\begin{subfigure}[b]{\textwidth}
	    \includegraphics[keepaspectratio,width=0.5\textwidth]{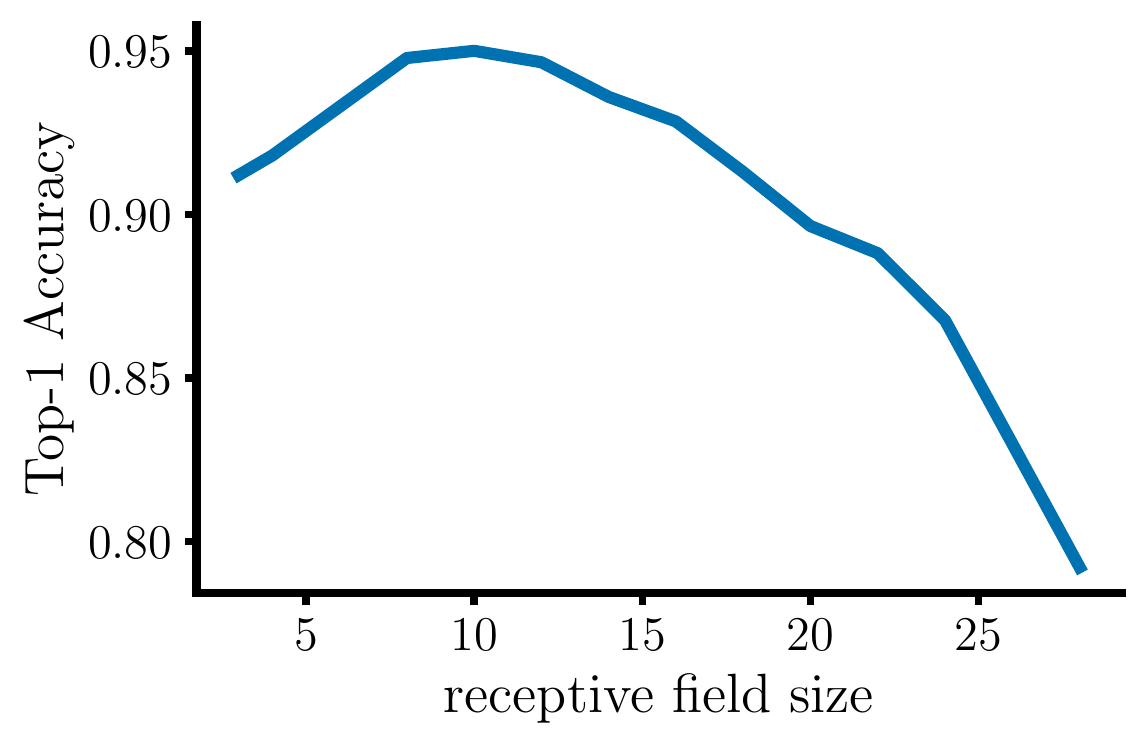}
	\end{subfigure}
	\begin{subfigure}[b]{\textwidth}
	    \includegraphics[keepaspectratio,width=0.5\textwidth]{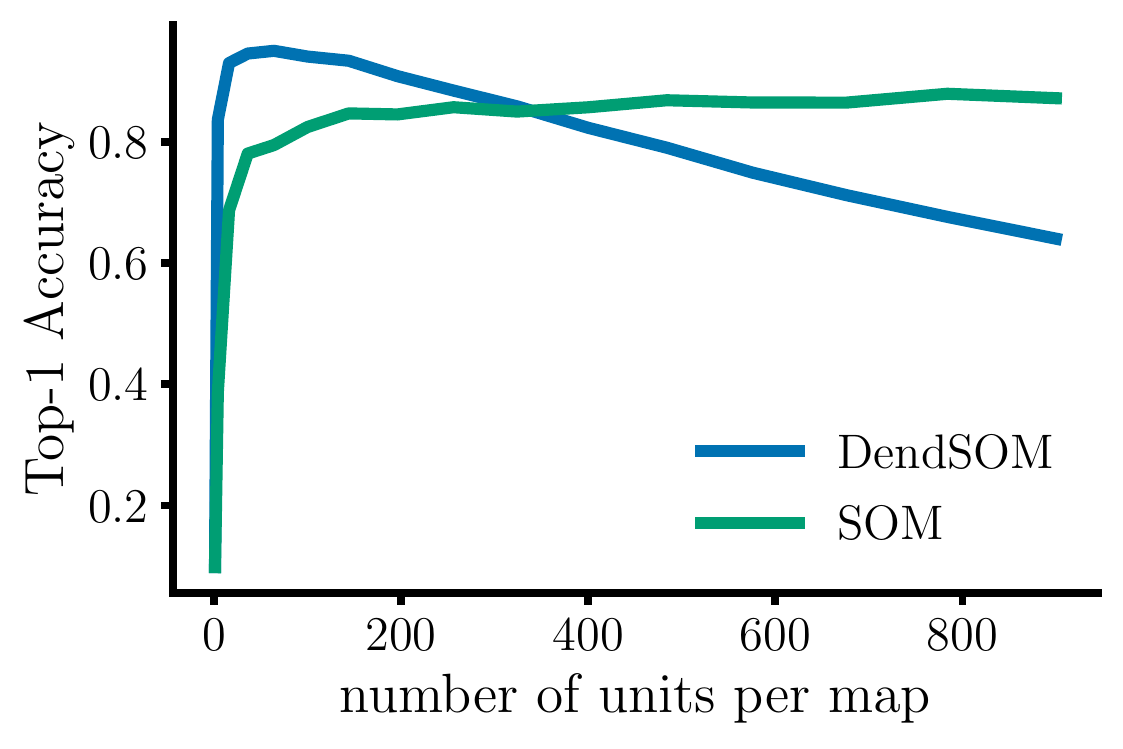}
	\end{subfigure}
	\caption{\textit{(top)} Receptive field size affects the DendSOM algorithm. The accuracy is a non-monotonic function of the receptive field size. The accuracy drops for small or large sizes, whereas for intermediate values, the model achieves its best performance. \textit{(bottom)} Influence of number of units per map over the DendSOM (blue) and SOM (green) algorithms.}
	\label{fig:rf_size}
\end{figure}

The neighborhood function (Figure \ref{fig:nf}) is a Gaussian function whose variance decreases over time due to the decrease of $\sigma$ function. Thus for $r_{exp}=2$, the variance of the $h$ Gaussian function doubles every $iter\_crit$ training steps, which allow for better mapping of the input space in the case of CL since the input space is not entirely presented to the model before the completion of the training process.

As shown in Figure \ref{fig:lr_nr_decay}, for $\lambda=1\text{e}3$, both $\alpha$ and $\sigma$ tend to zero after almost 1,000 time steps, and thus the DendSOM can not properly map the input space due to an insufficient number of input samples. However, for higher $\lambda$ values, the mapping process requires several epochs to be completed, which increases the training time of the algorithm. In the case of the DendSOM model, $\lambda=1\text{e}3$ is a good choice since incorporating the receptive fields allows for averaging in smaller regions of the input space and thus extracting simpler patterns.

What is more, Figure \ref{fig:ac_rx_cont} indicates that high $a_{crit}$ values tend to decrease the performance of the algorithm, while relatively low $r_{exp}$ values tend to improve the predictive ability of our model. Either high $r_{exp}$ or high $\alpha_{crit}$ values tend to increase both $\alpha$ and $\sigma$ dramatically, which in turn leads to weight vector overwriting. It is worth noting that the algorithm marks the highest accuracy score on MNIST and CIFAR-10 datasets for $\alpha_{crit}=0.005$ and $\alpha_{crit}=0.00005$, respectively.

Moreover, Figure \ref{fig:rf_size} (top) indicates that the accuracy score of our model increases for receptive field sizes smaller than $10$ since the patterns extracted from the input space are most likely to be informative patterns and thus non-typical patterns. On the other hand, when the receptive field size is greater than $10$, the DendSOM is reduced to a SOM, and therefore, the accuracy score decreases. It is worth to be mentioned that for receptive field size equal to $1$, the accuracy score is $9\%$.

Finally, Figure \ref{fig:rf_size} (bottom) shows that increasing the number of units improves the performance of the SOM algorithm. However, this does not seem to be the case for the DendSOM model since increasing the number of units above 100 decreases the model's predictive ability. This result is counter-intuitive, and thus, additional experiments have to be conducted to draw clear conclusions.

\bibliography{ref} 

\begin{thebibliography}{10}

\bibitem{kiela2020hateful}
D.~Kiela, H.~Firooz, A.~Mohan, V.~Goswami, A.~Singh, P.~Ringshia, and
  D.~Testuggine, ``The hateful memes challenge: Detecting hate speech in
  multimodal memes,'' {\em arXiv preprint arXiv:2005.04790}, 2020.

\bibitem{borji2014human}
A.~Borji and L.~Itti, ``Human vs. computer in scene and object recognition,''
  in {\em Proceedings of the IEEE conference on computer vision and pattern
  recognition}, pp.~113--120, 2014.

\bibitem{shen2019artificial}
J.~Shen, C.~J. Zhang, B.~Jiang, J.~Chen, J.~Song, Z.~Liu, Z.~He, S.~Y. Wong,
  P.-H. Fang, and W.-K. Ming, ``Artificial intelligence versus clinicians in
  disease diagnosis: systematic review,'' {\em JMIR medical informatics},
  vol.~7, no.~3, p.~e10010, 2019.

\bibitem{awasthi2019continual}
A.~Awasthi and S.~Sarawagi, ``Continual learning with neural networks: A
  review,'' in {\em Proceedings of the ACM India Joint International Conference
  on Data Science and Management of Data}, pp.~362--365, 2019.

\bibitem{stuart2016dendrites}
G.~Stuart, N.~Spruston, and M.~H{\"a}usser, {\em Dendrites}.
\newblock Oxford University Press, 2016.

\bibitem{branco2010single}
T.~Branco and M.~H{\"a}usser, ``The single dendritic branch as a fundamental
  functional unit in the nervous system,'' {\em Current opinion in
  neurobiology}, vol.~20, no.~4, pp.~494--502, 2010.

\bibitem{losonczy2008compartmentalized}
A.~Losonczy, J.~K. Makara, and J.~C. Magee, ``Compartmentalized dendritic
  plasticity and input feature storage in neurons,'' {\em Nature}, vol.~452,
  no.~7186, pp.~436--441, 2008.

\bibitem{london2005dendritic}
M.~London and M.~H{\"a}usser, ``Dendritic computation,'' {\em Annu. Rev.
  Neurosci.}, vol.~28, pp.~503--532, 2005.

\bibitem{spruston2008pyramidal}
N.~Spruston, ``Pyramidal neurons: dendritic structure and synaptic
  integration,'' {\em Nature Reviews Neuroscience}, vol.~9, no.~3,
  pp.~206--221, 2008.

\bibitem{chavlis2021drawing}
S.~Chavlis and P.~Poirazi, ``Drawing inspiration from biological dendrites to
  empower artificial neural networks,'' {\em Current Opinion in Neurobiology},
  vol.~70, pp.~1--10, 2021.

\bibitem{nguyen2019toward}
C.~V. Nguyen, A.~Achille, M.~Lam, T.~Hassner, V.~Mahadevan, and S.~Soatto,
  ``Toward understanding catastrophic forgetting in continual learning,'' {\em
  arXiv preprint arXiv:1908.01091}, 2019.

\bibitem{masse2018alleviating}
N.~Y. Masse, G.~D. Grant, and D.~J. Freedman, ``Alleviating catastrophic
  forgetting using context-dependent gating and synaptic stabilization,'' {\em
  Proceedings of the National Academy of Sciences}, vol.~115, no.~44,
  pp.~E10467--E10475, 2018.

\bibitem{li2017learning}
Z.~Li and D.~Hoiem, ``Learning without forgetting,'' {\em IEEE transactions on
  pattern analysis and machine intelligence}, vol.~40, no.~12, pp.~2935--2947,
  2017.

\bibitem{shin2017continual}
H.~Shin, J.~K. Lee, J.~Kim, and J.~Kim, ``Continual learning with deep
  generative replay,'' {\em arXiv preprint arXiv:1705.08690}, 2017.

\bibitem{kirkpatrick2017overcoming}
J.~Kirkpatrick, R.~Pascanu, N.~Rabinowitz, J.~Veness, G.~Desjardins, A.~A.
  Rusu, K.~Milan, J.~Quan, T.~Ramalho, A.~Grabska-Barwinska, {\em et~al.},
  ``Overcoming catastrophic forgetting in neural networks,'' {\em Proceedings
  of the national academy of sciences}, vol.~114, no.~13, pp.~3521--3526, 2017.

\bibitem{zenke2017continual}
F.~Zenke, B.~Poole, and S.~Ganguli, ``Continual learning through synaptic
  intelligence,'' in {\em International Conference on Machine Learning},
  pp.~3987--3995, PMLR, 2017.

\bibitem{maltoni2019continuous}
D.~Maltoni and V.~Lomonaco, ``Continuous learning in single-incremental-task
  scenarios,'' {\em Neural Networks}, vol.~116, pp.~56--73, 2019.

\bibitem{van2021class}
G.~M. van~de Ven, Z.~Li, and A.~S. Tolias, ``Class-incremental learning with
  generative classifiers,'' {\em arXiv preprint arXiv:2104.10093}, 2021.

\bibitem{bashivan2019continual}
P.~Bashivan, M.~Schrimpf, R.~Ajemian, I.~Rish, M.~Riemer, and Y.~Tu,
  ``Continual learning with self-organizing maps,'' {\em arXiv preprint
  arXiv:1904.09330}, 2019.

\bibitem{van2019three}
G.~M. Van~de Ven and A.~S. Tolias, ``Three scenarios for continual learning,''
  {\em arXiv preprint arXiv:1904.07734}, 2019.

\bibitem{kohonen2013essentials}
T.~Kohonen, ``Essentials of the self-organizing map,'' {\em Neural networks},
  vol.~37, pp.~52--65, 2013.

\bibitem{liu2015deep}
N.~Liu, J.~Wang, and Y.~Gong, ``Deep self-organizing map for visual
  classification,'' in {\em 2015 international joint conference on neural
  networks (IJCNN)}, pp.~1--6, IEEE, 2015.

\bibitem{aly2020deep}
S.~Aly and S.~Almotairi, ``Deep convolutional self-organizing map network for
  robust handwritten digit recognition,'' {\em IEEE Access}, vol.~8,
  pp.~107035--107045, 2020.

\bibitem{nakagawa2017classification}
A.~Nakagawa and A.~Kutics, ``Classification in big image datasets using
  layered-som,'' in {\em 2017 13th International Conference on Signal-Image
  Technology \& Internet-Based Systems (SITIS)}, pp.~143--150, IEEE, 2017.

\bibitem{cottrell2016theoretical}
M.~Cottrell, M.~Olteanu, F.~Rossi, and N.~Villa-Vialaneix, ``Theoretical and
  applied aspects of the self-organizing maps,'' in {\em Advances in
  self-organizing maps and learning vector quantization}, pp.~3--26, Springer,
  2016.

\bibitem{rolnick2018experience}
D.~Rolnick, A.~Ahuja, J.~Schwarz, T.~P. Lillicrap, and G.~Wayne, ``Experience
  replay for continual learning,'' {\em arXiv preprint arXiv:1811.11682}, 2018.

\bibitem{chaudhry2019continual}
A.~Chaudhry, M.~Rohrbach, M.~Elhoseiny, T.~Ajanthan, P.~K. Dokania, P.~H. Torr,
  and M.~Ranzato, ``Continual learning with tiny episodic memories,'' in {\em
  Proceedings of the 36th International Conference on Machine Learning}, Long
  Beach, California, PMLR 97, 2019.

\bibitem{mermillod2013stability}
M.~Mermillod, A.~Bugaiska, and P.~Bonin, ``The stability-plasticity dilemma:
  Investigating the continuum from catastrophic forgetting to age-limited
  learning effects,'' {\em Frontiers in psychology}, vol.~4, p.~504, 2013.

\bibitem{barros2018large}
R.~S.~M. Barros and S.~G. T.~C. Santos, ``A large-scale comparison of concept
  drift detectors,'' {\em Information Sciences}, vol.~451, pp.~348--370, 2018.

\bibitem{lecun1998mnist}
Y.~LeCun, ``The mnist database of handwritten digits,'' {\em http://yann.
  lecun. com/exdb/mnist/}, 1998.

\bibitem{xiao2017fashion}
H.~Xiao, K.~Rasul, and R.~Vollgraf, ``Fashion-mnist: a novel image dataset for
  benchmarking machine learning algorithms,'' {\em arXiv preprint
  arXiv:1708.07747}, 2017.

\bibitem{krizhevsky2009learning}
A.~Krizhevsky and G.~Hinton, {\em Learning multiple layers of features from
  tiny images}.
\newblock University of Toronto, 2009.

\end{thebibliography}

\bibliographystyle{ieeetr}
\end{document}